\documentclass[letterpaper, 10 pt, conference]{smc_ieeeconf}  

\IEEEoverridecommandlockouts                              

\usepackage{graphicx} 
\usepackage{amsmath} 
\usepackage{amssymb}  
\usepackage{bm,booktabs,algorithm,setspace}
\usepackage[caption=false,font=normalsize,labelfont=sf,textfont=sf]{subfig}
\usepackage{multirow}
\newcommand{\figintb}[1]{
\begin{minipage}[b]{0.14\columnwidth}
		\centering
		\raisebox{-.5\height}{\includegraphics[width=\linewidth]{#1}}
	\end{minipage}
}

\title{\LARGE \bf
Hierarchical Procedural Framework for Low-latency Robot-Assisted Hand-Object Interaction
}

\author{Mingqi Yuan$^{1,2*}$,
        Huijiang Wang$^{1*\dagger}$,~\IEEEmembership{Member,~IEEE,}  
        Kai-Fung Chu$^1$,~\IEEEmembership{Member,~IEEE,}  \\
        Fumiya Iida$^{1}$,~\IEEEmembership{Senior Member,~IEEE},
        Bo Li$^{2}$,
        Wenjun Zeng$^{3}$,~\IEEEmembership{Fellow,~IEEE}
\thanks{This work is supported, in part, by the Hong Kong SAR Research Grants Council (No. PolyU 15224823), the Department of Science and Technology of Guangdong Province (No. 2023A1515010592), the European Union's Horizon 2020 research, and the Innovation Programme under the Marie Skłodowska-Curie Grant (No. 101034337).}
\thanks{$^{1}$Bio-Inspired Robotics Lab, Department of Engineering, University of Cambridge, Cambridge CB2 1PZ, UK. $^{2}$Department of Computing, The Hong Kong Polytechnic University, Hong Kong SAR, CHN. $^{3}$Eastern Institute of Technology, Ningbo, Zhejiang, CHN.}
\thanks{$^*$These authors contributed equally to the work.}
\thanks{$^\dagger$Corresponding author: Huijiang Wang {\tt\small(hw567@cantab.ac.uk)}}
}

\begin{document}

\maketitle
\thispagestyle{empty}
\pagestyle{empty}

\begin{abstract}

Advances in robotics have been driving the development of human-robot interaction (HRI) technologies. However, accurately perceiving human actions and achieving adaptive control remains a challenge in facilitating seamless coordination between human and robotic movements. In this paper, we propose a hierarchical procedural framework to enable dynamic robot-assisted hand-object interaction (HOI). An open-loop hierarchy leverages the RGB-based 3D reconstruction of the human hand, based on which motion primitives have been designed to translate hand motions into robotic actions. The low-level coordination hierarchy fine-tunes the robot's action by using the continuously updated 3D hand models. Experimental validation demonstrates the effectiveness of the hierarchical control architecture. The adaptive coordination between human and robot behavior has achieved a delay of \(\le 0.3\) seconds in the tele-interaction scenario. A case study of ring-wearing tasks indicates the potential application of this work in assistive technologies such as healthcare and manufacturing. 

\end{abstract}


\section{INTRODUCTION}

The development of robotics and artificial intelligence (AI) has witnessed the revolution of the physical human-robot interaction (pHRI) \cite{soori2023artificial}. 
With the ability to detect and react with their human partners, these robots have been applied in a range of scenarios, including robot-assisted dressing \cite{zhang2022learning}, rehabilitation therapy \cite{ai2021machine}, medical treatment \cite{su2021physical}, and caregiving for the elderly and children \cite{wada2004effects,feil2011socially}.

To physically assist the human partners, these robots/agents are equipped with sensing and manipulation capabilities such that the co-bots can detect, interpret, and infer human cues, thereby providing tangible support to the human. To date, one typical task to achieve sophisticated pHRI has spotlighted the hand-object interaction (HOI), where humans manipulate objects and interact with the environment \cite{fan2021understanding, yang2022oakink}. Recent developments in computer vision have shed light on the robot-assisted HOI, such as using depth-enhanced vision systems and soft-robotic gloves to predict and assist with various hand postures for rehabilitation \cite{rho2024multiple}.

There are two fundamental technical challenges for robot-assisted HOI. One is hand pose estimation, which aims to accurately detect, track, and reconstruct the human hand in a dynamic environment under perturbation \cite{zimmermann2017learning}. Research has focused on 3D hand reconstruction using diverse technologies such as wearable sensors, radio frequency (RF) technologies, and acoustic methods \cite{wang2024towards}. In particular, the development of 3D pose estimation based on single RGB images has become prevalent in the current HOI community due to its ease of deployment. For instance, \cite{ge20193d} leveraged the graph convolutional neural network to reconstruct the whole 3D mesh of the hand surface from a single RGB image, which can well represent hand shape variations and capture local details. \cite{9710247} introduced a contact potential field approach, treating interactions as a spring-mass system to allow for dynamic adjustments based on contact physics, which enhances the physical plausibility of HOI predictions, even in cases of significant discrepancies in ground-truth data. In contrast, \cite{lin2021mesh} integrated the transformer architecture with graph convolution to create the MeshGraphormer, which adeptly captures both local and global interactions among mesh vertices and body joints. In summary, computer vision-based approaches provide a cost-effective and end-to-end solution for capturing intricate hand movements that advance the design of pHRI platforms.

\begin{figure*}[t!]
    \centering
    \includegraphics[width=\linewidth]{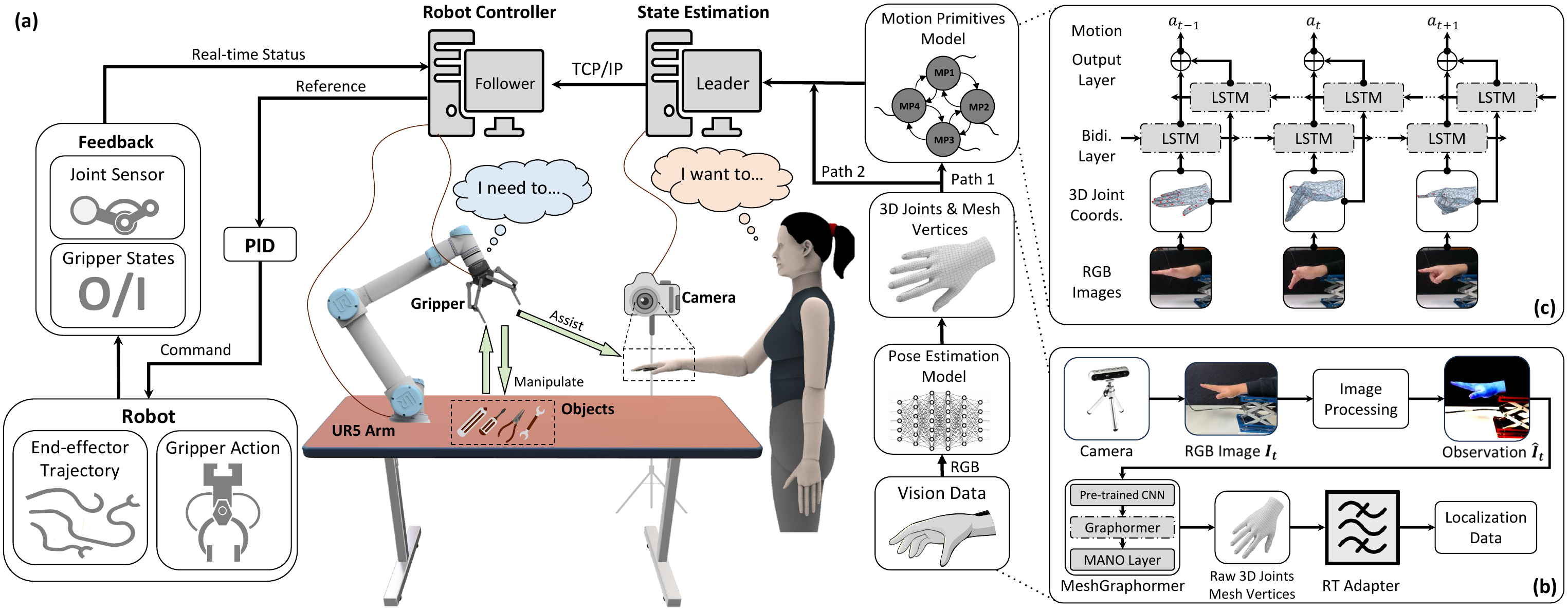}
    \vspace{-15pt}
    \caption{(a) Overview of the adaptive robotic system for robot-assisted HOI. (b) The module for RGB-based and 3D hand pose estimation. (c) The workflow of the proposed MPM, which is built on a bidirectional LSTM.}
    \label{fig:overview}
    \vspace{-15pt}
\end{figure*}

Another challenge lies in designing the robotic controller for robot-assisted manipulation. Various efforts have been put toward developing control interfaces, such as virtual joysticks \cite{campeau2018intuitive}, electromyography \cite{fall2016wireless}, inertial measurement units (IMUs) \cite{jain2015assistive}, eye gaze detection \cite{leroux2015combination} and brain-machine interfaces \cite{lampe2014brain}. Notably, motion primitive (MP)-based control has emerged as a promising method, which decomposes complex tasks into smaller, reusable actions that can be dynamically adapted. For instance, \cite{li2020skill} introduced a hierarchical skill-learning framework that combines dynamic MPs with Gaussian mixture models to learn motor skills from human demonstrations. This approach allows robots to encode complex, human-like motions and adapt to various interaction scenarios by reproducing these skills during cooperative tasks. Similarly, \cite{nguyen2022motion} developed an MP-based navigation planner that leverages deep collision prediction to ensure safe navigation in dynamic environments without relying on explicit maps. This method integrates MPs with deep learning techniques to predict collision probabilities, enabling real-time decision-making for navigation tasks.

In this paper, we propose a hierarchical procedural system for pHRI in the context of robot-assisted dressing. The first hierarchy consists of hand recognition and tele-manipulation. We proposed an MP-based robot controller to allow users to realize various customized motions. First, we introduce an RGB-based 3D hand pose estimation model to perform real-time vibration-tolerant state estimation. We trained a model to associate the hand state with MPs of different levels from a predefined MP library. Regarding the low-level hierarchy, we proposed a closed-loop control architecture to ensure the co-bot adjusts its motion to assist human partners adaptively. Finally, we introduce the action alignment model that allows for the adaptive orientation of the end-effector. Experimental tests have validated the effectiveness of the proposed hierarchical control framework in HOI tasks. Our proposed human-robot collaborative model contributes to the future development of robot-assisted healthcare and medical treatment services.

\section{RGB-BASED 3D HAND POSE SENSING}\label{sec:handPose}
A key challenge in achieving efficient and robust robot-assisted HOI is accurately sensing and modeling the human hand in real time. This requires capturing the 3D coordinates of hand joints and surface vertices, which are crucial for hand localization and providing feedback during robot actions. In this paper, we leverage these 3D coordinates for human intention recognition and adaptive robot control, using a transformer-based algorithm, MeshGraphormer \cite{lin2021mesh}, for 3D hand perception from RGB images. We selected MeshGraphormer for its (i) high modeling accuracy, particularly in capturing intricate hand poses and subtle joint movements, which is essential for precise control in complex interactions; (ii) real-time detection capabilities; and (iii) robust generalization across diverse environments due to its training on large-scale datasets, making it highly resilient to occlusions and lighting variations.

As shown in Figure~\ref{fig:overview}(b), MeshGraphormer first leverages a pre-trained CNN to extract grid features and a global feature vector from the given RGB images. These features are then tokenized and fed to a multi-layer transformer-based encoder to output coarse mesh tokens, and a multi-layer perceptron (MLP) is utilized to perform upsampling to obtain full mesh tokens and mesh vertices. Finally, the 3D coordinates of hand joints are computed using a MANO model from \cite{MANO:SIGGRAPHASIA:2017}. Denote by $F_{\bm\theta}$ the model of MeshGraphormer with network parameters $\bm{\theta}$, we have $(\bm{x},\bm{y},\bm{z})=F_{\bm\theta}\left(P(\bm{I})\right)$, where $\bm{I}$ is a raw RGB image, $\bm{x},\bm{y},\bm{z}$ are 3D coordinates vectors, and $P(\cdot)$ denotes the image preprocessing operation including resizing, cropping, normalizing, etc.


In practice, data oscillation may occur due to slight hand shaking during real-time detection, leading to unstable localization. To address the problem, we leverage a moving average filter to smooth the detection data. Additionally, we implement asynchronous processing pipelines to optimize our system's responsiveness and minimize latency in real-time applications for image preprocessing and pose estimation. Moreover, the hand pose estimation model can be replaced by a body pose model in our system, allowing the system to be adapted for full-body interaction scenarios. This flexibility is essential in complex environments where understanding hand and body gestures can significantly enhance the robot's ability to interpret human intentions accurately.

\section{MOTION PRIMITIVE MODEL}   \label{sec:PM_model}
Equipped with the derived 3D hand information, we further propose a motion primitive model (MPM) that converts human hand motions into robotic actions in real time. By utilizing the 3D coordinates information of hand joints, our model interprets and translates specific hand motions into corresponding mechanical responses. In particular, we define a set of MPs of different levels, where low-level primitives handle basic actions such as grasping or releasing, which are directly mapped to robot actions without additional inputs. High-level primitives, on the other hand, process more complex gestures that may involve multiple sequential or parallel actions, integrating contextual data from the environment or other sensory feedback. This hierarchical structure allows the MPM to adaptively handle a wide range of human intentions, ensuring fluid, context-aware collaboration between the human and robot in dynamic tasks.

We build the MPM using a bidirectional long short-term memory (LSTM) network \cite{hochreiter1997long}, a type of recurrent neural network particularly suited for sequence prediction problems. This choice is important as it enables the model to retain context over time, which is essential for interpreting continuous and dynamic hand gestures. The LSTM network is trained on a self-collected dataset of hand motion sequences to encompass a spectrum of potential gestures and their corresponding robotic actions. Through this training, the network learns the individual gestures and their transitions, ensuring a fluid and intuitive control flow. Denote by $M_{\bm{\phi}}$ the MPM with network parameters $\bm{\phi}$, we have
\begin{equation}
    \bm{a}_{t}=M_{\bm{\phi}}\left(\{{\rm Concat}(\bm{x}_{i},\bm{y}_{i},\bm{z}_{i})\}_{i=t-N+1}^{t}\right),
\end{equation}
where $\bm{a}_t$ is the predicted probability distribution of motions, $a_{t}=\underset{i}{\rm argmax}\,\bm{a}_{t}[i]$ will be the output motion class, and $N$ is a fixed sequence length. Meanwhile, the cross-entropy \cite{cover1999elements} is selected as the loss function to train the MPM:
\begin{equation}
    L(\bm\phi)=-\sum_{i=1}^{T}\sum_{k=1}^{K}\hat{\bm{a}}_{t}[k]\log(\bm{a}_{t}[k]),
\end{equation}
where $T$ is the number of time steps in the sequence, $K$ is the number of possible actions, and $\hat{\bm{a}}_{t}[k]$ is the ground truth indicator for class $k$ at time $t$ (1 if the class label $k$ is the correct classification for $t$, and 0 otherwise).


In real-time detection, since the time it takes to complete a human hand motion is uncertain, the prediction results will change suddenly during the motion. In order to ensure the credibility of the recognition results, we use the following filter:
\begin{equation}
    a_{t}=\begin{cases}
    a_{t},\quad &\mathrm{Var}(A)=0,|A|=N \\
    \inf,\quad &\mathrm{else}
    \end{cases}
\end{equation}
where $A$ is a queue of length $N$ that stores the real-time results from MPM, $|A|$ is the number of stored results, and $\mathrm{Var(\cdot)}$ is the variance (0 when all the results are identical).

Finally, we would like to highlight the advantages of the proposed MPM. First, MPM directly uses the 3D coordinate sequences of hand joint points as input, which maximizes the difference between different gestures in the feature space and effectively improves the detection accuracy and robustness. Moreover, MPM requires minimal training data and exhibits strong generalization capabilities. For example, a model trained on data from a single human hand can successfully recognize gestures from unknown human hands, as shown in the experiment section. Additionally, this system can be easily extended to a wide range of tasks by designing additional motion primitives or combining existing ones at different levels. After binding them to specific hand gestures, the model can be quickly trained and adapted for various applications, enabling efficient and intuitive robot control in diverse scenarios. This approach ensures flexibility and scalability, making the system suitable for a broad spectrum of HRI tasks.

\section{ROBOT CONTROL}\label{sec:controller}
\subsection{Experimental Setup}

\begin{figure}[t]
    \centering
    \includegraphics[width=\columnwidth]{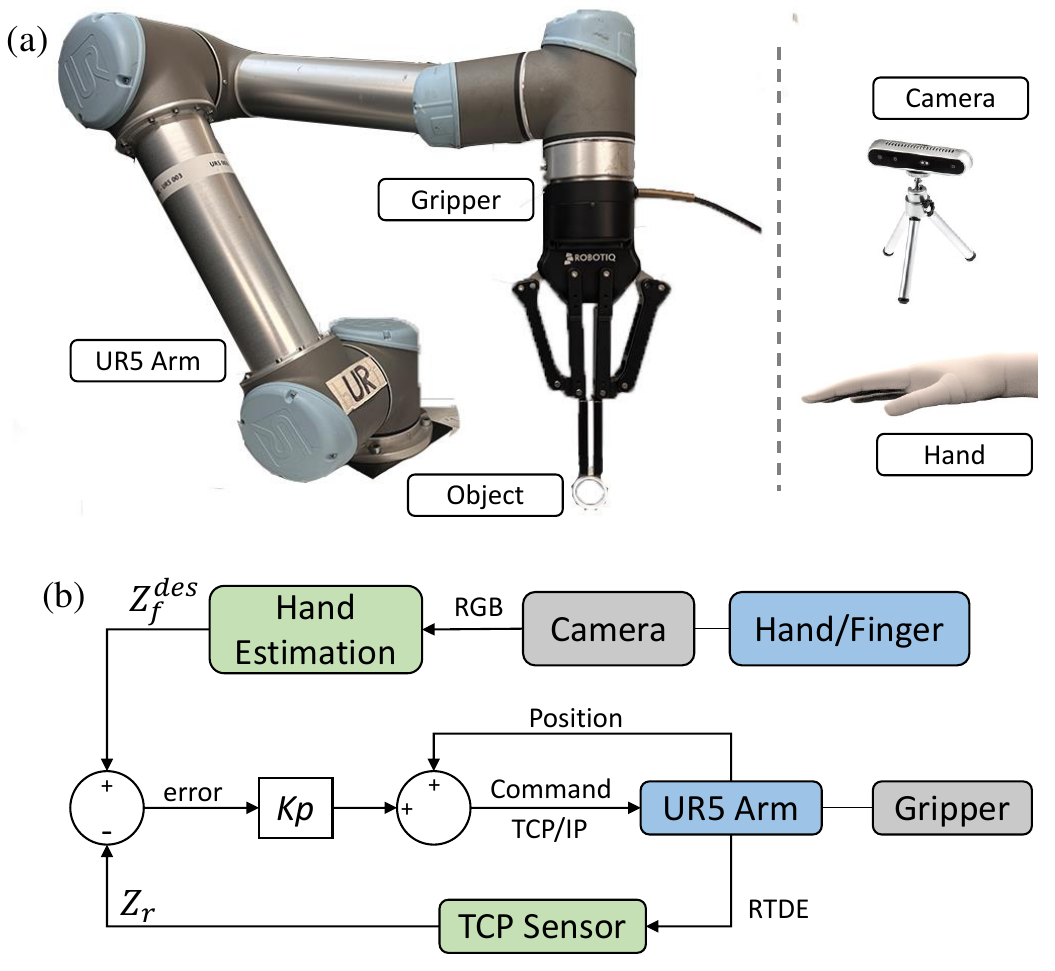}
    \vspace{-25pt}
    \caption{The systematic framework of the hand-object interaction. (a) The robot setup includes a UR5 robotic arm equipped with a gripper for grasping objects. (b) The overall control architecture.}
    \label{fig:robotSetup}
    \vspace{-10pt}
\end{figure}

We utilize the UR5 manipulator as the robotic arm and a two-finger RobotiQ gripper as the end-effector to execute pick-and-place tasks. Hand pose recognition is facilitated by an RGB camera (Intel RealSense). The experimental setup is illustrated in Figure \ref{fig:robotSetup}(a).

To control the UR5, we employed the real-time data exchange (RTDE) control interface to enable real-time manipulation of the tool center point (TCP) of the robot arm. Motor servoing was achieved using the \textit{servoL} function at a control rate of 500 Hz. As a safety precaution during implementation, we configured a safety zone around the end-effector to mitigate the risk of potential human injury.

\subsection{Controller Architecture}

The interaction between the hand and the object requires the robot to exhibit adaptive behaviors based on human input, necessitating hierarchical control over the robot arm. The gripper control operates as an open-loop system, enabling the gripper to execute pinching and releasing actions on the object. The UR5 TCP control relies on predefined motion primitives. Upon program initialization, the camera can capture hand motion and identify these primitives. When a motion primitive is detected, the robot executes predetermined incremental motions, allowing the user to adjust the spatial location of the TCP manually.

Closed-loop control occurs during the cooperation stage (see Figure \ref{fig:robotSetup}(b)). When the cooperation condition is triggered, the robot continuously approaches the human hand at a constant speed while tracking hand pose and receiving hand state information, such as finger angular deviation and fingertip location. Utilizing this data, the robot dynamically adjusts its vertical position and orientation to ensure successful object delivery. Throughout this process, the robot adapts both its spatial position and end-effector orientation to collaborate effectively with the human operator. The transition between these control modes is triggered by the detection of specific motion primitives, {\em i.e.}, the "Ring" primitive, signaling the robot to enter a cooperation mode. The entire control architecture is depicted in Figure \ref{fig:robotSetup}.

\subsection{Action Alignment}       \label{sec:actionAlign}
A crucial aspect is the coordination between the object and the human. When the gripper holds the object, the gripper-object system is considered a rigid body. When the gripper moves towards the human finger, the controller dynamically adjusts to accommodate finger bending and environmental perturbations. We design a motion alignment model to enable real-time adjustment of the robot's end-effector pose.

\begin{figure}[h!]
    \centering
    \includegraphics[width=\columnwidth]{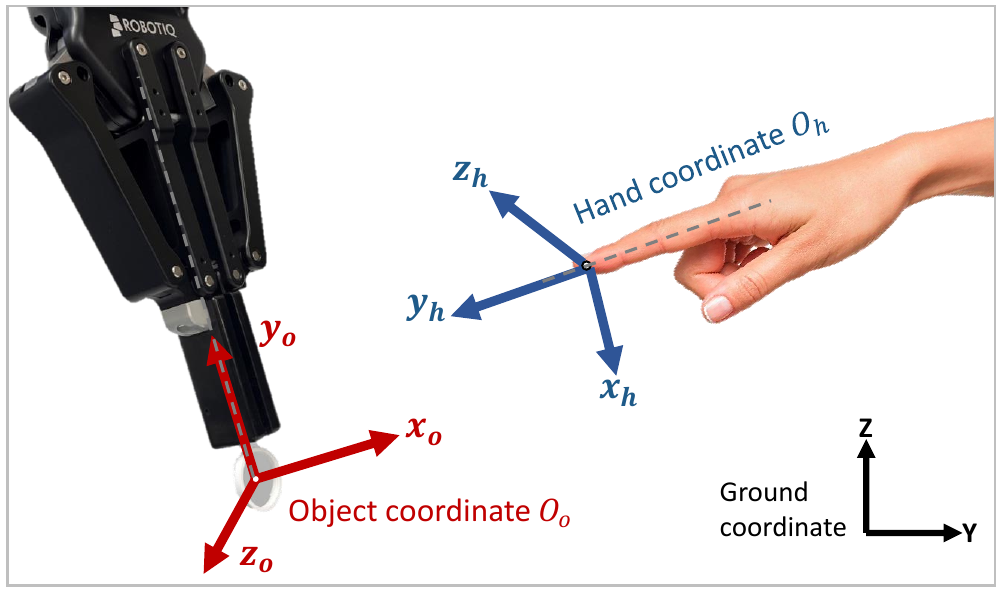}
    \vspace{-20pt}
    \caption{The spatial coordinates alignment between the hand and the object.}
    \label{fig:alignment}
\end{figure}

In Figure~\ref{fig:alignment}, we define the spatial coordinate of the object as \(O_o = [X_o, Y_o, Z_o]\). The center is established at the object's center, with the x-axis aligned along the object's normal direction. The y-axis corresponds to the gripper's central line, while the z-axis follows the right-hand rule. Similarly, with the employed hand pose recognition model, we determine the fingertip's coordinate as \(O_h = [X_h, Y_h, Z_h]\). The hand coordinate system originates from the fingertip, with the y-axis aligned with the fingertip's direction. The x-axis is positioned perpendicular to the y-axis and towards the bending side. The z-axis is set based on the right-hand rule. For the two coordinate systems defined in this initial system, it is always possible to find a rotation matrix $\mathrm{\textbf{T}}$ that performs the transformation from \(O_o\) to the \(O_h\):

\begin{equation}
 O_h = \mathrm{\textbf{T}} \cdot O_o    .
\end{equation}

In order to align the object with the finger, the object-based body coordinate should be aligned accordingly. For the given pointing pose in Figure \ref{fig:alignment}, to ensure the aligned orientation, we first compute the spatial angular deviations as:

\begin{equation}
\begin{matrix}
\alpha = \cos^{-1}\left\langle x_o, y_h \right\rangle, \beta = \cos^{-1}\left\langle y_o, z_h \right\rangle.
\end{matrix}
\end{equation}

To ensure that the ring can be worn on the finger, the orientation should follow:
\begin{equation}
    x^o_g - x^h_g= 0,y^h_g - y^o_g= v_h\cdot t,z^h_g - z^o_g= v_v\cdot t
\end{equation}
The subscript indicates the coordinate system being referenced, while the superscript indicates the object (ring or finger). The symbols \( v_h \) and \( v_v \) represent the robot's velocity in the horizontal and vertical directions, respectively. To ensure the alignment, we need to have \(\alpha = -1\) and \(\beta = 1\) such that the robot’s TCP synchronizes with the human hand's vertical movements, and when finger bending occurs, the TCP adjusts to align the object with the finger's central axis.

\section{RESULTS}\label{sec:result}
\subsection{Hand Recognition Performance}

\begin{table}[h!]
\vspace{-10pt}
\caption{Defined motion primitives for the ring-wearing task.}
\label{tb:mp}
\centering
\begin{tabular}{ccccc}
\toprule
\textbf{Motion} &  Keep &  Come&  Back& Ring \\
\midrule
\textbf{\begin{tabular}[c]{@{}c@{}}RGB\\ Image\end{tabular}}   &  \figintb{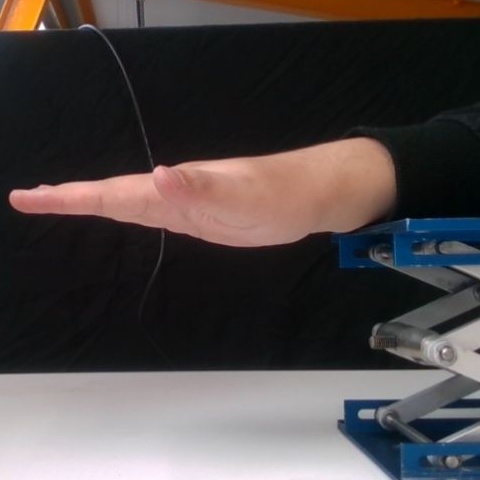} & \figintb{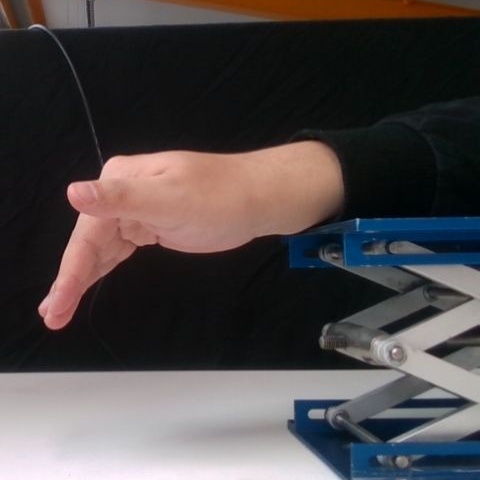} & \figintb{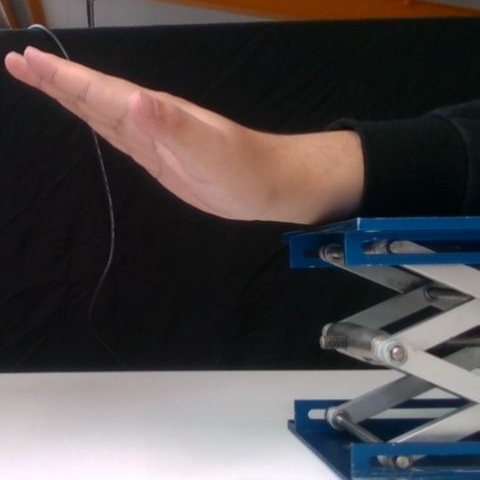} & \figintb{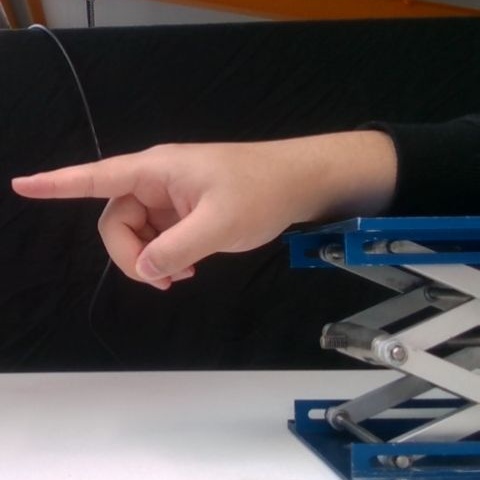} \\
\bottomrule
\end{tabular}
\vspace{-20pt}
\end{table}

\begin{figure}[h!]
    \centering
    \includegraphics[width=\linewidth]{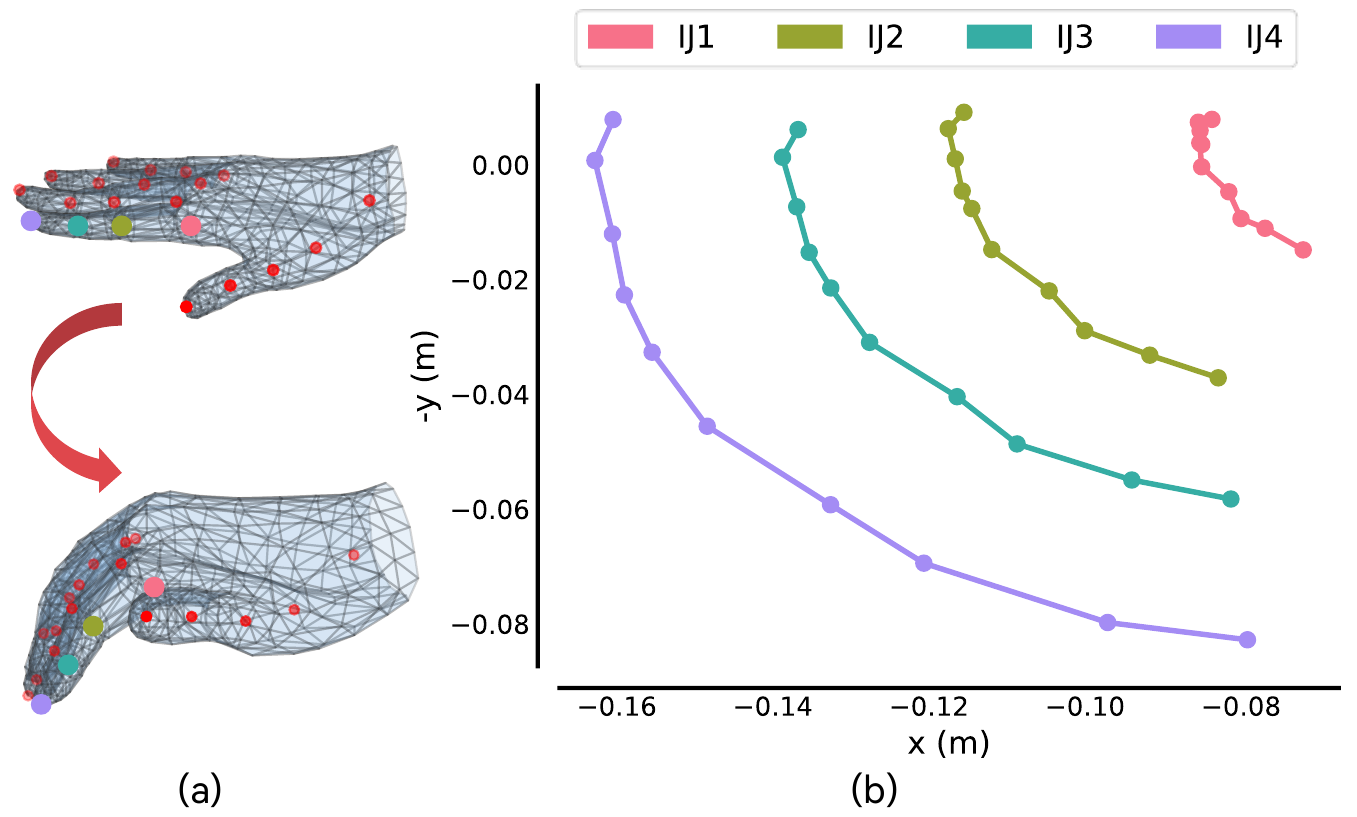}
    \vspace{-20pt}
    \caption{3D modeling of the process of converting from the motion "Keep" to the motion "Come". (a) Mesh reconstruction of the hand motion, in which the filled circles represent the joints. (b) Changes in the coordinates of joints during the motion conversion process. Here, "IJ1-4" denotes the four joints of the index finger.}
    \label{fig:motion_change}
\end{figure}

\begin{figure*}[h!]
    \centering
    \includegraphics[width=0.95\linewidth]{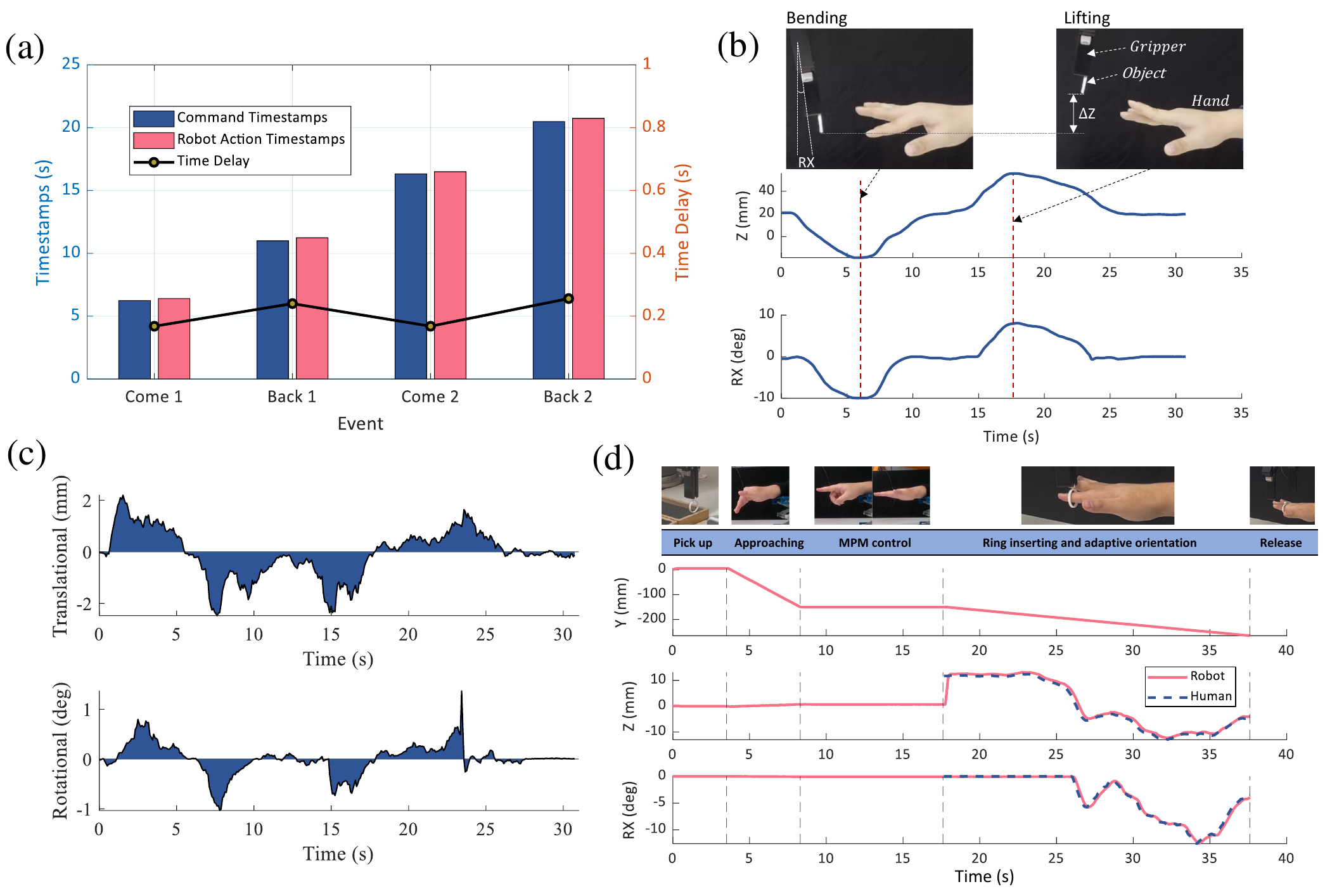}
    \vspace{-10pt}
    \caption{Real-world performance of the proposed tele-interaction task. (a) Comparison of the timestamps for human hand commands and corresponding robot actions. (b) Adaptive orientation of the UR5 robot arm based on human hand input. (c) Translational and rotational errors between the predicted and actual TCP (Tool Center Point) positions. (d) Multiple stages of the robot-assisted dressing task are demonstrated in the ring-wearing scenario.}
    \label{fig:realTimePerformance}
    \vspace{-15pt}
\end{figure*}

We first evaluate the effectiveness of the designed 3D hand pose module. Table~\ref{tb:mp} illustrates the pre-defined motion primitives while Figure~\ref{fig:motion_change} demonstrates the 3D modeling of the transition process between two distinct hand motions. In the "Keep" motion, the y-axis coordinates of the four joints remain relatively unchanged, while under the "Come" motion, the coordinates of these joints progress according to the hand motion. These observations validate the capture and model of the hand motion changes, providing reliable reference information for the subsequent robot control.

\subsection{MPM Training}
\subsubsection{Data Preparation}
For each motion, the human participants were instructed to make and maintain the motion, and then collect 2000 consecutive frames of RGB image data, and the subject made slight motion adjustments during the collection process. Then, these frames were used to extract 3D coordinates of 21 hand joints using the aforementioned pose estimation model. For a single trajectory, we sampled every 10 frames and intercepted a sequence of length 10. As a result, a motion dataset with 1000 samples was obtained, and the data shape of each input is $(10, 63)$. Furthermore, we use stratified sampling to divide the dataset into a training set (80\%) and a validation set (20\%) to ensure that each type of action has the same number of samples.
\subsubsection{Model Training}
For the MPM, we leveraged a bidirectional LSTM with 2 recurrent layers, and the input size is (10, 63), and the number of features in the hidden state is 64. The LSTM is followed by a fully connected layer that maps the LSTM output to 4 classes corresponding to different hand motions and a softmax function to output the probability distribution over these classes. Finally, we trained the MPM for 100 epochs using a learning rate of 2.5e-4 and a batch size of 512.

\subsection{Adaptive Pose Adjustment}

\subsubsection{Motion Primitives Transition}

Each motion primitive is activated individually. Supporting Video S1 demonstrates the motion control using the MPM, where the UR5 robot’s movements along the Y-axis are primarily controlled by two MPs. The robot exhibits a strong dynamic response to these commands, with different MPs triggering actions such as forward and backward movements. Figure \ref{fig:realTimePerformance}(a) illustrates the time delays between the robot's actions and the human commands during two consecutive come-back MPM commands. It can be seen that at each timestamp, the robot consistently lags behind the human command. For the 'come' action, the delay is 0.168 seconds, while the 'back' MPM action experiences a delay of approximately 0.24 seconds. The maximum control delay remains under 0.3 seconds, demonstrating the cobot's effective command-tracking capabilities within the control architecture.

\subsubsection{Adaptive Orientation}

The robot handles environmental and human-induced disturbances in real-world implementations. Figure \ref{fig:realTimePerformance}(b) shows the robot's Z-axis translational and X-axis angular displacement trajectories. Over 30 seconds, the human bends and then lifts the index finger. The robot maintains its grip on the object (ring) and adjusts its TCP orientation in terms of vertical height and pointing angle. The robot achieves a maximum displacement of \(\Delta Z\) of 74.23 mm and a rotational angle of 18.0 degrees. These adjustments enable the robot to maintain a suitable position for dressing, validating the effectiveness of the adaptive controller for action alignment discussed in Section \ref{sec:actionAlign}. The error between the predicted TCP and the actual translational TCP is shown in Figure \ref{fig:realTimePerformance}(c), where the errors fluctuate around zero over time. The maximum absolute translational error is 2.48 mm, while the maximum absolute rotational error reaches 1.37 degrees.

\subsection{Robot-assisted Ring Wearing}

Figure \ref{fig:realTimePerformance}(d) presents the results of the robot-assisted ring-wearing task (see supporting Video S3). The first and second rows of curves represent the TCP's motions along the Y and Z axes, respectively, while the third row shows the angular displacement around the x-axis, measured with respect to the ground coordinate system shown in Figure \ref{fig:alignment}. In the open-loop hierarchy, the robot first performs a pre-programmed action to pick up the ring. It is then controlled by the MPM model, enabling the human operator to guide the end-effector's approach using the 'Come' hand pose. Once the operator specifies the desired location, they activate the 'Ring' motion primitive, shifting the robot from passive control to active cooperation. From 17.6 seconds onward, the robot operates within the closed-loop hierarchy, continuously adjusting its movement towards the hand and dynamically compensating for any perturbations caused by the human hand. The dashed blue curve represents the motion of the CV-based hand pose estimation. As the human adjusts their finger position or location, the robot adapts in real-time, ensuring the ring remains properly oriented until it is successfully delivered. At 37.5 seconds, the robot completes the ring-wearing task by releasing the ring, thereby demonstrating the effectiveness of the robot in real-world applications such as robot-assisted dressing.

\section{CONCLUSION}\label{sec:conclusion}
In this paper, we developed a hierarchical procedural framework for low-latency robot-assisted hand-object interaction. We presented an open-loop hierarchy that leverages learning-based state estimation for real-time 3D reconstruction of hand pose. Meanwhile, a LSTM-based model translates these poses into motion primitives, which are then executed as pre-defined robotic actions in a tele-manipulation manner. Regarding the low-level coordination hierarchy, we have designed a closed-loop controller that enables the spatial alignment between the end-effector and the fingering, ensuring that the object is manipulated collaboratively. Experimental tests have shown that the system exhibits a time delay of \(\leq 0.3\) seconds, a translational error of \(\leq 2.5\) mm, and a rotational error of \(\leq 1.5\) degrees.

In the case study of a ring-wearing task, the adaptive robotic system demonstrates its effectiveness in assisting its human teammate and its robustness against environmental perturbations. Future work can be developed to expand the motion primitive sets and refine interaction dynamics, enhancing the applications of pHRI systems in practical environments.

\bibliography{ref}
\bibliographystyle{ieeetr}

\end{document}